\documentclass{article}[12pt]
\usepackage[english]{babel}
\usepackage{amsthm}
\usepackage{soul}
\usepackage{xcolor}
\usepackage{graphicx}
\usepackage{algorithm}
\usepackage{caption}
\usepackage{algpseudocode}
\usepackage[margin=1in]{geometry}
\usepackage{array, booktabs, caption}
\usepackage{ragged2e}
\usepackage{makecell}
\usepackage[utf8]{inputenc}
\usepackage[T1]{fontenc}    
\usepackage{mathptmx}       
\usepackage{microtype}      

\usepackage{amsmath}
\usepackage{amssymb}
\usepackage{amsthm}

\usepackage[margin=1in]{geometry}
\usepackage{titlesec}
\usepackage{parskip}        
\usepackage{enumitem}
\usepackage{setspace}       
\usepackage{hyperref}
\usepackage{graphicx}
\usepackage{xcolor}

\newtheorem{theorem}{Theorem}[section]
\newtheorem{definition}{Definition}[section]
\newenvironment{proof_doc}{\paragraph{Proof:}}{\hfill$\square$}

\title{\Large Some Complexity Results for Robustness Verification of Binarized Neural Networks}
\author{\textbf{Harshit Goyal, Sudakshina Dutta}\\
Indian Institute of Technology Goa} 
\date{}

\newtheorem{lemma}{Lemma}[section]
\newtheorem{corollary}{Corollary}

\begin{document}

\maketitle

\section{Introduction}
This paper focuses on \emph{Binarized Neural Networks} (BNNs)\cite{bnn}, a class of neural networks whose activations and, in many cases, weights are restricted to binary values. BNNs are attractive because of their computational efficiency and suitability for deployment on resource-constrained hardware. At the same time, their discrete structure makes them particularly amenable to formal reasoning and complexity-theoretic analysis. We know that proving satistisfiability of a property for a DNN is NP-hard\cite{Reluplex}. Understanding the computational difficulty of verification problems for BNNs is therefore important from both theoretical and practical perspectives.

The primary objective of this paper is to investigate the computational complexity of three verification problems for BNNs. First, we study the satisfiability problem of linear properties for BNNs, namely whether there exists an input assignment that causes a given network to produce a designated output. We show that this problem is NP-complete by establishing membership in NP and providing a polynomial-time reduction from the boolean satisfiability problem\cite{ref_sat}. Second, we analyze robustness under non-uniform image occlusion and third, we investigate robustness under uniform image occlusion. Unlike general adversarial perturbations, uniform occlusion\cite{occrob} constrains all pixels within the occluded region to assume the same value and non-uniform occlusion\cite{occrob} constrains all pixels to assume values within a range of values. We demonstrate that this restriction induces a piecewise-constant behavior in the network output as a function of the occlusion color, which leads to a polynomial-time robustness-checking algorithm.

To support these results, the next section introduces the necessary background on neural network classifiers, binarized neural networks, image representations, occlusion models, robustness, and computational complexity. Subsequently, we present the formal problem definitions, complexity-theoretic proofs, and robustness analysis that form the core contributions of this paper.

\section{Background}
This section introduces the fundamental concepts and definitions required for the problems studied in this project, including neural network classifiers, binarized neural networks, occlusion models, robustness, and relevant computational complexity classes.

\subsection{Neural Network Classifiers}\label{subsec:nn_classifier}
\begin{figure}
    \centering
    \includegraphics[width=0.5\linewidth]{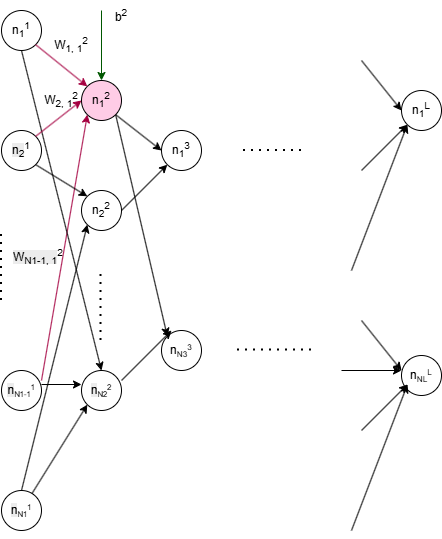}
    \caption{An example neural network}
    \label{fig:nn_example}
\end{figure}

A pre-trained neural network classifier is a function
$f:\mathbb{R}^{N_1} \rightarrow \{1,2,\dots, N_L\}$ that maps an input vector (e.g., an image) with a dimension $N_1$ to one of $N_L$ discrete class labels as shown in Figure ~\ref{fig:nn_example}. More precisely, $N_i$ is the number of neurons in the layer $i$. We consider feedforward neural networks with $L$ layers. For the first neuron (with pink colour) of the second layer the pre-activation and post-activation values are defined as:
\begin{align*}
    z^{2}_1 &= W^{2}_{1, 1} n_{1}^1 + W^{2}_{2, 1} n_{2}^1 + W^{2}_{N_1-1, 1} n_{N_1-1}^1 + b^{2}, \\
    n_{1}^2 &= \sigma(z_1^{2}),
\end{align*}
Note that the $k^{th}$ neuron of layer $j$ is denoted by $n_{k}^j$; the weight on an edge connecting $i^{th}$ neuron to $k^{th}$ neuron in the $j^{th}$ layer is denoted by $W_{i, k}^{j}$; the bias on layer $j$ is $b^{j}$ (we assume that it has the same values for all neurons in a layer).
For any neuron $n_{k}^j$, $2\leq j \leq L-1$ the pre-activation and post-activation values are defined by the following equations.
\begin{align*}
    z^{j}_{k} &= \sum_i W^{j}_{i, k}*n_{i}^{j-1} + b^{j}, \\
    n_{k}^j &= \sigma(z_k^{j}),
\end{align*}
We have used $\sigma$ to be $sign$ activation function for the present work. The output layer produces $y^{(L)}$ which is implemented by ARGMAX activation function. More precisely, $n_{j}^L = max_i(n_{i}^{L-1})$, where the neuron $i$ in the layer $L-1$ is connected to a neuron $j$ in the layer $L$. Note that the computation of $z_k^j$ as given above can be represented by matrix multiplication and addition.

\subsection{Binarized Neural Networks (BNNs)}\label{subsec:BNN}
A Binarized Neural Network (BNN) is a neural network in which the weights, activations, or both are restricted to binary values, typically $\{-1, +1\}$. In this project, we focus on BNNs with $\mathrm{sign}$ activation functions applied to all hidden layers.
The activation function $\sigma$ is defined as:
$$\sigma(v) = \mathrm{sign}(v) = \begin{cases} +1, & v \ge 0, \\ -1, & v < 0. \end{cases}$$

\subsection{Image Representation}
An image is represented as a vector
$$x \in \mathbb{R}^n,$$
where each component corresponds to a pixel intensity. For an image of height $h$ and width $w$, we have $n = h \times w$. This vectorized representation allows the image to be used directly as input to the neural network.

\subsection{Occlusion}\label{sec:occ}
Occlusion is a form of input perturbation where a region of the image is modified or hidden.
Let $x$ = [$x_1$ $x_2$ $\cdots$ $x_n$] be an input image and let $O \subseteq \{1, \dots, n\}$ denote the set of pixel indices covered by an occlusion window. In this work, we consider that the pixels in $O$ constitute a rectangular region. The position  of the occlusion region is denoted by the coordinate of the top-left corner  pixel of $O$. 

\subsubsection*{Non-Uniform Occlusion}\label{subsubsec:non_uni_occ}
In non-uniform occlusion, each pixel in the occluded region can be modified independently. Consider an unoccluded image pixel to be $x_i$, $1 \leq i \leq n$,  which is occluded (i.e., $i \in O$); $x_i'$ is image pixel after occlusion. We consider $L_{\infty}$-norm based occlusion i.e., maximum absolute value of an image pixel is bounded by a fixed value, say $\epsilon$; hence, $-\epsilon \leq x'_i \leq \epsilon$. Pixels outside $O$ remain unchanged. This model allows maximal flexibility and represents a strong adversarial perturbation.

\subsubsection*{Uniform Occlusion}
\label{subsubsec:uni_occ}In uniform occlusion, all occluded pixels take the same value $c$. Formally,
$$x'_i = \begin{cases} c, & i \in O, \\ x_i, & i \notin O. \end{cases}$$
This restriction significantly limits the space of possible perturbations and plays a key role in reducing computational complexity.

\subsection{Robustness}
Consider the set of images generated by applying a perturbation $\mathcal{P}$ on an image $x$ to be $\mathcal{P}(x)$ (A specific perturbation e.g., occlusion can modify an image in different ways if the occlusion positions, size, etc change). A neural network classifier $f$ is said to be robust if the predicted label of $x$ does not change for any element in the set $\mathcal{P}(x)$. Formally,
$\forall x' \in \mathcal{P}(x), \quad f(x') = f(x).$
The robustness decision problem asks whether such robustness holds for a given input, network, and perturbation function.

\section{Computational Complexity Results}

\subsection{Problem 1: Satisfiability of Linear Properties for BNNs}

\begin{definition}[BNN-SAT]
Consider a pre-trained Binarized Neural Network ($\mathrm{BNN}$) $\Psi$ which has two output classes. In other words, $\Psi(x) = +1$ or $\Psi(x) = -1$ on input vector $x$. The $\mathrm{BNN}$-$\mathrm{SAT}$ problem asks whether there exists an input assignment $x \in \{-1,1\}^n$ such that $\Psi(x) = +1$.
\end{definition}\label{def:BNN_SAT}

\subsubsection*{Membership in NP}

A nondeterministic algorithm can guess an input assignment $x \in \{-1,1\}^n$ and evaluate the network $\Psi(x)$ in polynomial time. Since the evaluation of a single neuron involves only matrix multiplication, addition followed by the $\mathrm{sign}$ activation (as discussed in ~\ref{subsec:nn_classifier}) (a constant number of arithmetic and activation steps per layer), the overall verification process is polynomial.
Thus, $\mathbf{BNN}$-$\mathbf{SAT} \in \mathbf{NP}$.

\subsubsection*{NP-Hardness Proof}

To prove $\mathbf{NP}$-hardness, we reduce from the boolean Satisfiability problem ($\mathbf{SAT}$).

Let $\varphi$ be a Conjunctive Normal Form ($\mathrm{CNF}$) formula:
\[
\varphi = C_1 \wedge C_2 \wedge \dots \wedge C_m,
\]
where each clause $C_j$ is a disjunction of literals over variables $x_1, x_2, \dots, x_k$. Note that a literal can be $x_i$ or $\neg x_i$. 

\paragraph{Reduction Construction}
We construct a BNN $\Psi$ that outputs $+1$ if and only if $\varphi$ is satisfiable, by mapping the boolean values $\mathrm{True/False}$ to the BNN activation values $+1/-1$, respectively.

\begin{enumerate}[label=\textbf{Step \arabic*:}, leftmargin=2em]
\item \textbf{Variable Gadgets:}
Each boolean variable is directly represented by an input neuron $x_i \in \{-1,1\}$.

\item \textbf{NOT Gadget :}
The literal $\neg x_i$ is implemented by a gadget with two neurons; the input neuron corresponds to the value $x_i$ and the output neuron $y$ outputs $-x_i$. The weight corresponding to the edge connecting the neurons is $-1$ and the bias for the output neuron is 0.
\begin{center}
\includegraphics[width=0.4\textwidth]{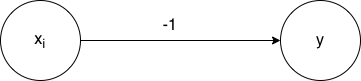}
\end{center}

\item \textbf{OR Gadget :}
The OR gadget corresponding to any disjunctive clause $(x_1 \vee x_2 \vee x_3)$ is constructed using five neurons. The inputs constitute four neurons, where three come from $x_1$, $x_2$, $x_3$ and the fourth neuron assumes the value 2. These four neurons connect to the only neuron in the next layer with edges and the weights on all edges are $+1$; the only neuron in the second layer generates the output. 
\[
y = \operatorname{sign}\!\left(\sum_{i=1}^3 x_i + 2\right).
\]
\begin{center}
\includegraphics[width=0.35\textwidth]{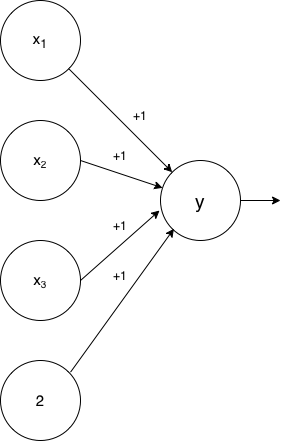}
\end{center}
Note that $y = -1$ only if $x_1$ = $x_2$ = $x_3$ = $-1$. Otherwise, $y = +1$.

\item \textbf{AND Gadget (Formula Satisfaction):}
The AND gadget corresponds to conjunction of $m$ clauses $C_1$, $C_2$, $\cdots$ $C_m$ and it is constructed using $m+2$ neurons. The inputs constitute four neurons, where three come from $C_1$, $C_2$, $\cdots$, $C_m$ and the $m^{th}$ neuron assumes the value $(m-1)$. The first $m$ neurons connect to the only neuron in the next layer with edges and the weights on all edges are $+1$; the last neuron connects the only neuron in the second layer in the next layer with an edge and the weight on the edge is $-1$; the neuron in the second layer generates the output. 
\[
y = \operatorname{sign}\!\left(\sum_{i=1}^m x_i - (m-1)\right).
\]
\begin{center}
\includegraphics[width=0.35\textwidth]{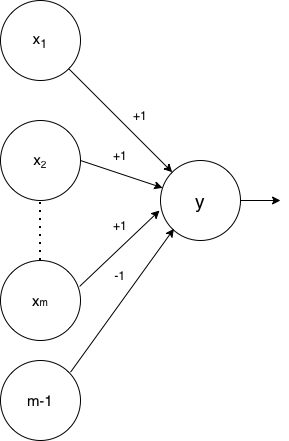}
\end{center}
Note that $y = -1$ only if $x_1$ = $x_2$ = $\cdots$ = $x_m$ = $+1$. Otherwise, $y = -1$.
    
\end{enumerate}

\paragraph{Correctness of Reduction}
The network $\Psi$ outputs $1$ if and only if each clause neuron output $y_j=1$, which occurs exactly when all clauses of $\varphi$ are satisfied (True). Therefore,
\[
\Psi(x) = 1 \iff \varphi(x) = \text{True}.
\]
Thus, $\Psi$ is satisfiable if and only if $\varphi$ is satisfiable.

\paragraph{Complexity}
The construction of $\Psi$ from $\varphi$ involves a number of neurons and connections are linear in the size of $\varphi$, hence the reduction can be performed in polynomial time.

\begin{theorem}
Since $\mathbf{BNN}$-$\mathbf{SAT}$ is both in $\mathbf{NP}$ and $\mathbf{NP}$-hard (via reduction from $\mathbf{SAT}$), it follows that $\mathbf{BNN}$-$\mathbf{SAT}$ is $\mathbf{NP}$-Complete.
\end{theorem}

\subsection{Problem 2: Robustness under Non-Uniform Occlusion}

\begin{definition}[Occlusion Robustness Decision Problem]
Consider a pre-trained Binarized Neural Network ($\mathrm{BNN}$) $\Psi$ which has two output classes. In other words, $\Psi(x) = +1$ or $\Psi(x) = -1$ on input vector $x$. Let $I \in \mathbb{R}^{m \times n}$ be an input image and $\Psi$ to  be a trained BNN classifier. Let $w, h \in \mathbb{N}$ define the width and height of a rectangular occlusion window.

An occlusion is defined by:
\begin{enumerate}
    \item A placement $(i,j)$ such that the window $W_{i,j, w, h} \subseteq \{1,\dots,m\} \times \{1,\dots,n\}$ is an axis-aligned $w \times h$ subregion of the image $I$.
    \item An assignment $\delta$ of values to the pixels in $W_{i,j, w, h}$, allowing arbitrary (non-uniform) values within the occlusion window in the range $[0, 255]$.
\end{enumerate}

Let $I'$ denote the image obtained from $I$ by replacing the pixels in $W_{i,j, w, h}$ following $\delta$. The occlusion robustness decision problem asks whether there exist a window position $(i,j)$ and an occlusion assignment $\delta$ such that:
\[
\Psi(I') \neq \Psi(I)
\]
If such a position and assignment exist, the network is said to be \textit{not robust} to non-uniform occlusions of size $w \times h$; otherwise, it is robust to all such occlusions.
\end{definition}

\subsubsection*{Proof of Membership in NP}

We show that the decision problem belongs to the complexity class $\mathbf{NP}$ by exhibiting a polynomial-size certificate and a polynomial-time verification algorithm.

\paragraph{Certificate}
For a ``yes'' instance of the problem, there exists an input vector $\mathbf{c} = (c_1, c_2, \dots, c_n), c_i \in \{-1, +1\}$ such that the BNN classifier $\Psi$ outputs the positive label, i.e., $\Psi(\mathbf{c}) = +1$. We choose this input vector $\mathbf{c}$ as the certificate. The size of the certificate is $O(n)$, which is polynomial in the size of the input layer of the network.

\paragraph{Verifier Algorithm}
Given an instance of the problem and a certificate $\mathbf{c}$, the verifier $V$ proceeds as follows:
\begin{enumerate}
    \item \textbf{Forward Evaluation:} Perform a single forward pass through the neural network $\Psi$ using $\mathbf{c}$ as the input.
    \item \textbf{Neuron Activation Computation:} For each neuron $k$, from the first hidden layer up to the output layer, compute its activation according to:
    \[
    y_k = \operatorname{sign}\left(\sum_i w_{k,i} y_i + b_k\right)
    \]
    where $y_i$ denotes the activations from the previous layer (or the input components $c_i$ for the first hidden layer), and $w_{k,i}$ and $b_k$ are the corresponding weights and bias.
    \item \textbf{Output Check:} Let $y_{\text{out}}$ denote the activation of the output neuron.
    \item \textbf{Acceptance Condition:} If $y_{\text{out}} = +1$, the verifier accepts. If $y_{\text{out}} = -1$, the verifier rejects.
\end{enumerate}

The verifier runs in time polynomial in the size of the network, since it performs a single forward pass involving a polynomial number of neurons and arithmetic operations. Therefore, the verification procedure runs in polynomial time. Since there exists a polynomial-size certificate that can be verified in polynomial time for every ``yes'' instance, the problem belongs to $\mathbf{NP}$.

\subsubsection*{NP-Hardness Proof}

We reduce from boolean SAT.

\paragraph{SAT Instance}
Let $\varphi = C_1 \wedge C_2 \wedge \dots \wedge C_m$ be a CNF formula over variables $x_1, x_2, \dots, x_q$
\footnote{Note that each clause $C_i$ of $\varphi$, $1 \leq i \leq m$, may itself be a disjunction of linear constraints, e.g.,
$a_1x_1 + a_2x_2 + \cdots + a_px_p \;\Box\; b$, where $\Box \in \{<,\leq,>,\geq\}$. Assume that each integer variable is represented in binary using $w=\lceil \log_2(U+1)\rceil$ bits, where $U$ is an upper bound on its value. Multiplication by constant coefficients, addition of the resulting binary numbers, and comparison with the constant $b$ can be implemented by polynomial-size Boolean circuits. In particular, multiplication by a constant and comparison require $O(w)$ gates, while the sum of $p$ $w$-bit integers can be computed using $O(pw)$ gates. Therefore, the resulting arithmetic circuit has size $O(pw)$. By Tseitin's transformation~\cite{sat}, a Boolean circuit of size $s$ can be translated into an equisatisfiable CNF formula with $O(s)$ auxiliary variables and $O(s)$ clauses. Hence, every such linear constraint can be reduced to an equisatisfiable SAT formula in polynomial time.}

\paragraph{Reduction Construction}
We construct an instance of the occlusion robustness problem. We assume that for the original image $I$, $\Psi(I) = -1$.

\textbf{Step 1: Occluded Image construction}\\
An flattened occluded image $I'$ = \{$p_1$, $p_2$, $\cdots$, $p_q$\} $\in$ $\mathbb{R}^q$ is constructed from the SAT assignment on $\varPhi$. We assume that the occlusion window cover the entire image. A pixel assumes the color value 255 if True, 0 otherwise. 

\textbf{Step 2: Preprocessing layer}\\
The first layer of the constructed BNN converts $I'$ to a binary occluded image using a threshold function. 
For each pixel $p_i$, the corresponding preprocessing neuron computes

\[
z_i=p_i-\tau,
\]

where $\tau=127.5$ is the threshold. The neuron output is given by

\[
b_i=
\operatorname{sign}(z_i)
=
\begin{cases}
-1, & p_i<\tau,\\
+1, & p_i\ge \tau.
\end{cases}
\]

Thus, every pixel is interpreted as a boolean variable, where

\[
p_i<127.5
\Longleftrightarrow
b_i=-1,
\]

and

\[
p_i\ge127.5
\Longleftrightarrow
b_i=+1.
\]

The binary outputs
$\{b_1,b_2,\ldots,b_{q}\}$
are subsequently used as the inputs to the remaining BNN, which is
constructed exactly as in the SAT-to-BNN-SAT reduction.

\textbf{Step 3: BNN Construction} \\
We construct a BNN classifier $\Psi: \mathbb{R}^{q} \to \{-1, +1\}$ that computes $\Psi(I')$. The network outputs $+1$ iff the encoded assignment satisfies $\varphi$. Such boolean functions are representable by BNNs of polynomial size.

\paragraph{Correctness of the Reduction}

We prove that
\[
\varphi \text{ is satisfiable }
\iff
\text{the constructed BNN is not robust under non-uniform occlusion}.
\]

\noindent
($\Rightarrow$)
Assume that $\varphi$ is satisfiable.
Let
\[
\alpha:\{x_1,\ldots,x_q\}\rightarrow\{True, False\}
\]
be a satisfying assignment.

Construct an occluded image $I'$ by assigning each pixel as

\[
p_i=
\begin{cases}
0,&\alpha(x_i)=False,\\
255,&\alpha(x_i)=True.
\end{cases}
\]

By construction, the preprocessing layer maps these pixel values to the
corresponding binary values by applying a thresholding function

\[
b_i=sign(p_i - \tau), \qquad 1\le i\le q.
\]

Hence the first hidden layer receives exactly the satisfying assignment
$\alpha$.

Since the remaining BNN is identical to the SAT-to-BNN-SAT construction,
the network evaluates the boolean formula $\varphi$.
As $\alpha$ satisfies $\varphi$, the output neuron evaluates to $+1$.
Since the original image is classified as $-1$, the occluded image is
misclassified.
Therefore the constructed BNN is not robust under non-uniform occlusion.

\medskip

\noindent
($\Leftarrow$)
Assume that the constructed BNN is not robust.
Then there exists an occluded image $I'$ whose pixel values $p_i$ lie in
$[0,255]$ and for which

\[
\Psi(I')\neq\Psi(I).
\]

The preprocessing layer converts the pixel values of $I'$ into the binary
values

\[
b_1,b_2,\ldots,b_q\in\{-1, +1\}.
\]

These values induce the boolean assignment
$\alpha(x_i)=True$ for $b_i = +1$, $\alpha(x_i)=False$, otherwise ($1 \leq i \leq q$.

Since BNN-SAT has resulted a misclassification, the boolean formula $\varphi$ evaluates to true under the assignment $\alpha$.

Hence $\alpha$ satisfies $\varphi$.
Therefore $\varphi$ is satisfiable.

Thus,

\[
\varphi \text{ is satisfiable }
\iff
\text{the constructed BNN is not robust under non-uniform occlusion}.
\]

\begin{theorem}
Since non-uniform robustness verification (NUOP) is both in $\mathbf{NP}$ and $\mathbf{NP}$-hard (via reduction from $\mathbf{SAT}$), it follows that $\mathbf{NUOP}-\mathbf{SAT}$ is $\mathbf{NP}$-Complete.
\end{theorem}
\subsection{Problem 3: Robustness under Uniform Occlusion}

\begin{theorem}
The problem of deciding if a BNN $\Psi$ (Definition ~\ref{subsec:BNN}) classifies an input image $x$, uniformly occluded with colour $c$ (section ~\ref{subsubsec:uni_occ}), to the correct label is solvable in polynomial time with respect to the input $x$ and the total number of network parameters.
\end{theorem}\label{th:uni_robustness}

\subsubsection*{Proof of Polynomial-Time Solvability}

The main challenge is that the color for uniform occlusion, $c$, can assume values from $\mathbb{R}$. We shall see that $\Psi(x)$ on uniformly occluded image $x$ is a piecewise function of the colour $c$. 

\begin{lemma}[First-Layer Pre-Activation is Affine]
For a fixed occlusion position $(a, b)$ (as mentioned in section ~\ref{sec:occ}), the pre-activation value $z_{k}^{2}$ of $k^{th}$ neuron in the first hidden layer is an affine function of the colouring constant $c$.
\end{lemma}
\begin{proof_doc}
The pre-activation value for the $k^{th}$ neuron in the first hidden layer is 
$z^{2}_{k}$ = $\sum_i W^{2}_{i, k}*n_{i}^1$ + $b^{2}$ (as mentioned in section ~\ref{subsec:nn_classifier}). For the proof, we assume the existence of an original unoccluded image 
(referred to as the \emph{base image}) and a corresponding mask image that specifies the occlusion region. The mask image is assumed to be binary: a pixel value is $1$ if the corresponding pixel belongs to the occlusion region $O$, and $0$ otherwise.

The pixel values of the original image are treated as the base values. The occluded image is then computed using both the base image and the mask. Specifically, the $k^{\text{th}}$ input pixel is defined as $n_{k}^1$ = $(1 - n_{k, mask}^1)$ $\cdot$ $n_{k, base}^1$ + $c \cdot n_{k, mask}^1$,
where $c$ denotes the constant occlusion value.

Thus, pixels outside the occlusion region retain their base values, 
while pixels inside the occlusion region are replaced by the constant value $c$.

Now the pre-activation value $z^{2}_{k}$ = $\sum_i W^{2}_{i, k}\cdot n_{i}^1$ + $b^{2}$ = $\sum_i W^{2}_{i, k} \cdot n_{i}^1$ + $b^{2}$ = $\sum_i W^{2}_{i, k}\cdot((1 - n_{i, mask}^1) \cdot n_{i, base}^1
+ c \cdot n_{i, mask}^1)$ + $b^{2}$ = $m_k\cdot c + q_k$. Both $m_{k}$ and $q_{k}$ are constants depending only on the network parameters.
\end{proof_doc}
\begin{definition}(Breakpoints)
Let $f : \mathbb{R} \rightarrow \mathbb{R}$ be a piecewise-defined function. A point $x_0 \in \mathbb{R}$ is called a \emph{breakpoint} of $f$ if either
\begin{itemize}
    \item the functional form of $f$ changes at $x_0$, or
    \item the derivative of $f$ is discontinuous at $x_0$.
\end{itemize}
Formally, $x_0$ is a breakpoint if there exist linear functions $f_1$ and $f_2$ such that
\[
f(x) =
\begin{cases}
f_1(x), & x < x_0, \\
f_2(x), & x > x_0,
\end{cases}
\]
where $f_1 \neq f_2$.
\end{definition}
Note that we use sign activation function for BNN (as mentioned in ~\ref{subsec:BNN}) and it causes breakpoints in the values of post-activation function.
\begin{lemma}[First-Layer Post-Activation is Piecewise-Constant]
The post-activation $n_{k}^2$ = $\sigma(z_k^{2})$ is a piecewise-constant function (as mentioned in section ~\ref{subsec:BNN}) of $c$. If $m_{k}\ne0$, this function has at most one breakpoint at $c_{k} =-q_{k}/m_{k}$.
\end{lemma}\label{lem:breakpoints}

\begin{lemma}[Propagation of Piecewise-Constant Functions]
For any hidden layer $l \in [1, N_L - 1]$ and any neuron $k$ in that layer, the post-activation output $n_{k}^l$ is a piecewise-constant function of $c$. Moreover, all breakpoints are generated in the first hidden layer, and no additional breakpoints are introduced in the subsequent layers.
\end{lemma}
\begin{proof_doc}
Consider an occluded input image to be $x$ with occlusion position  $(a, b)$. The image $x$ is subjected to a pre-trained BNN with $L$ layers. The proof proceeds by induction on the layer depth $l$.
\begin{itemize}
    \item \textbf{Base Case ($l=1$):} Proven by Lemma ~\ref{lem:breakpoints}. The set of all breakpoints of first hidden layer is demoted as $B_2$.
    \item \textbf{Inductive Step:} Assume the lemma holds for layer $l-1$ i.e., for any node $n_{i}^{l-1}$, only breakpoints from the set $B_2$ are used. The pre-activation value for the neuron $k$ in the layer $l$ is $z_{k}^{l}$; it is a linear combination of the piecewise-constant functions $n_{i}^{l-1}$, if the neuron $i$ is connected to neuron $k$ in the layer $l$. A linear combination of such functions is also piecewise-constant, with breakpoints contained in the same set $B_2$. Applying the $\mathrm{sign}$ activation to a piecewise-constant function cannot introduce new breakpoints, as the function can only change value at points where its input changes sign, which can occur only at existing breakpoints.
\end{itemize}
\end{proof_doc}

\begin{corollary}[Final Logits are Piecewise-Constant]
The final network outputs (logits) $z_{k}^{L}$ for all $k\in\{1,...,N_L\}$ are piecewise-constant functions of $c$, with all breakpoints contained in $B_2$.
\end{corollary}

\paragraph{The Algorithm}
We discuss the pseudocode to determine uniform occlusion robustness for a BNN $\Psi$ with an input image uniformly occluded with the colour $c$.

\begin{enumerate}
    \item \textbf{Initialize:} Compute the correct classification using the original, unoccluded image $x_{orig}$, $y_{orig}=\Psi(x_{orig})$.
    \item \textbf{Iterate Positions:} Let the set of all valid occlusion position be $\mathcal{M}$.\\
    For each position $(a, b) \in \mathcal{M}$ :
    \begin{enumerate}
        \item \textbf{Compute Breakpoints:} Compute the first-layer breakpoint set $B_2$. The size $|B_2| \le N_{2}$. The breakpoints partition the domain into $|B_2|+1$ open intervals. The set $C_{test}$ includes reprentative values from each interval. The piecewise behavior of the network is the same for every interval.
        \item \textbf{Create Test Set:} 
          Sort the breakpoints. Let the sorted set be $\{b_1, b_2, \cdots, b_{N_2}\}$. The intervals are \{$[-\infty, b_1)$, $[b_1, b_2)$, $[b_2, b_3)$, $\cdots$ $[b_{N_2 - 1}, b_{N_2})$, $[b_{N_2}, \infty)$\}. Except the first interval, representative values of each interval $[b_i, b_j)$ can be included by including $b_i$ in $C_{test}$. To include the initial open interval, we can construct a finite test set
        \[
            C_{\text{test}} = B_2 \cup \{b_1-0.1\},
        \]
        by selecting one representative value from each of these intervals. Consequently, the cardinality of $C_{\text{test}}$ is $N_2 + 1$.\\
        \textbf{Example} : 
        Let \( B_2 = \{-1, -1.5\} \). These breakpoints partition the real line into the intervals (see the example ~\ref{ex:uni_occ})
        \[
        (-\infty, -1.5), \quad [-1.5, -1), \quad [-1, \infty).
        \]
        The piecewise behavior of the network changes precisely at the points \( -1 \) and \( -1.5 \).

        In particular, for the first interval $(-\infty, -1.5)$, consider the breakpoint \( -1.5 \) for the function \( F = 2c + 3 \) in the first hidden layer. Then
        \[
            \operatorname{sign}(F) =
            \begin{cases}
                1, & \text{if } F \ge 0 \; (\text{i.e., } c \ge -1.5), \\[6pt]
                -1, & \text{if } F < 0 \; (\text{i.e., } c < -1.5).
            \end{cases}
        \]
        Thus, the sign of \(F\) changes at \( c = -1.5 \). Similarly, the sign of the function $G = 2c + 2$ also changes at $c = -1$. Therefore, the breakpoint $-1.5$ serves as a representative for the interval $[-1.5, -1)$, while the breakpoint $-1$ represents the interval $[-1, \infty)$.

        To obtain a representative for the first interval $(-\infty, -1.5)$, an additional test point slightly less than $-1.5$, namely $-1.5 - 0.1 = -1.6$, is included.
        \item \textbf{Test Values:} For each test value $c\in C_{test}$:
        \begin{itemize}
            \item Construct the perturbed input $x$.
            \item Calculate the network's prediction $y_{pred}=\Psi(x)$.
            \item \textbf{Check for Misclassification:} If $y_{pred}\ne y_{orig}$, return NON-ROBUST.
        \end{itemize}
    \end{enumerate}
    \item \textbf{Final Result:} If all positions and test values are checked without finding a misclassification, return ROBUST.
\end{enumerate}

\paragraph{Complexity Analysis}
The number of occlusion positions $|\mathcal{M}|$ is $O(m \times n)$, where $m$ and $n$ are width and height of the image $x_{orig}$, respectively. The complexity for finding robustness for a single position is:
$$O(\text{Breakpoint Computation} + Sorting + |C_{test}| \cdot \text{Forward Pass})$$
$$O((N_1\cdot N_2) + N_2logN_2 + N_2.(N_1\cdot N_2 \cdots N_{L-1})) \text{}$$
The total runtime is $O(|\mathcal{M}|\cdot ((N_1\cdot N_2) + N_2\cdot logN_2 + N_2\cdot(N_1\cdot N_2 \cdots N_{L-1}))$. 
This complexity is polynomial in the input image size   and network parameters, proving the claim in Theorem ~\ref{th:uni_robustness}.

\begin{algorithm}
\caption{Checking Uniform Occlusion Robusness of a BNN}
\begin{algorithmic}[1]
\Require $\Psi$ be a pretrained BNN defined as a sequence of layers $\{l_1, l_2, \ldots, l_L\}$, where $l_1$ denotes the input layer and $|l_i| = N_i$ for $1 \leq i \leq L$; with each $i$, the layer $l_i$, $1 \leq i \leq l_L -1 $ performs an affine transformation parameterized by the weight matrix $W^i$ and bias vector $b^i$,  followed by applications of sign activation functions with the penultimate layer $l_{L-1}$ is an affine layer, and the final layer 
$l_L$ applies the $\mathrm{ARGMAX}$ activation function. Let $x \in \mathbb{R}^{m \times n}$ be an MNIST image with a uniform occlusion 
of color $c$, where $m \cdot n = N_1$, and let the ground-truth label of the original image, $x_{orig}$ be $y_{orig}$.

\Ensure To output ``robust" if the BNN is occlusion robust, ``non-robust" if it is not robust
\State Initialize $O_{l_1, sign} = x_f$, where $x_f$ is the flattened vector of the image $x_{}$ of size $(m * n) \times 1$
\For{each layer $j$ in the set of all layers hidden layers $L = \{l_2, \cdots, l_{L-1}\}$}
\State Compute the weighted sum $O_{l_{j}}$ = $O_{l_{j-1}, sign} \cdot W^{(j)} + b^{(j)}$ 
    \State Apply the sign activation $O_{l_{j}, sign} = sign(O_{l_{j}})$
\EndFor 
\State Compute the network output by applying $\mathrm{ARGMAX}$ to the output of $l_{L-1}$:
\[
O_{l_L} \gets \mathrm{ARGMAX}(O_{l_{L-1}}).
\]
\State Let $B_2$ = \{$b_1$, $b_2$, $\cdots$, $b_{N_2}$\} be the set of breakpoints induced by layer $l_2$. The number of breakpoints satisfies $|B_2| \leq N_2$. 
\State Compute $C_{test} = B_2 \cup \{b_1 - 0.1\}$
\For{each region $c_i$ in $C_{test}$}
    \If{$O_{l_L} \neq y_{orig}$}
        \State \Return ``non-robust''
    \EndIf
\EndFor
\State \Return ``robust''

\end{algorithmic}
\end{algorithm}
\newpage

\subsection{Example: Piecewise Constant Behavior under Uniform Occlusion}\label{ex:uni_occ}

To illustrate the theoretical result that the network output is a piecewise constant function of the occlusion color $c$, we present a concrete numerical example.

\subsubsection{Setup}
Consider a $3 \times 2$ input image (6 pixels) flattened into a vector $\mathbf{x} \in \mathbb{R}^6$.
\begin{itemize}
    \item \textbf{Original Image:} The pixel values are given by the vector $\mathbf{x}_{orig} = \begin{bmatrix} 2 & 3 \\ 13 & 14\\ 15 & 16 \end{bmatrix}$.
    \item \textbf{Occlusion:} A $2 \times 2$ occlusion window is placed at position $(2,1)$ (covering the last 4 pixels).
    \item \textbf{Perturbed Input:} For a coloring constant $c \in \mathbb{R}$, the perturbed input vector becomes:
    \[
    \mathbf{x} = \begin{bmatrix} 2 & 3 \\ c & c\\ c & c \end{bmatrix}
    \]
\end{itemize}

\subsubsection{Neural Network Architecture}
We use a 4-layer feedforward network with the following parameters. The activation function is $\sigma(v) = \operatorname{sign}(v)$.

\begin{itemize}
    \item \textbf{Layer 1 ($6 \to 2$):}
    \[
    W^2 = \begin{bmatrix} 1 & -1 & -1 & 1 & 1 & 1 \\ -1 & 1 & -1 & 1 & 1 & 1 \end{bmatrix}, \quad b^{2} = \begin{bmatrix} 3 \\ 2 \end{bmatrix}
    \]

    \item \textbf{Layer 2 ($2 \to 3$):}
    \[
    W^3 = \begin{bmatrix} 1 & 1 \\ -1 & 1 \\ -1 & -1 \end{bmatrix}, \quad b^{3} = \begin{bmatrix} 1 \\ 2 \\ 1 \end{bmatrix}
    \]

    \item \textbf{Layer 3 ($3 \to 2$):}
    \[
    W^4 = \begin{bmatrix} 1 & -1 & 1 \\ 1 & 1 & 1 \end{bmatrix}, \quad b^{4} = \begin{bmatrix} 1 \\ -2 \end{bmatrix}
    \]

    \item \textbf{Layer 4 (Output, $2 \to 4$):}
    \[
    W^5 = \begin{bmatrix} 1 & 1 \\ -1 & 1 \\ 1 & -1 \\ -1 & -1 \end{bmatrix}, \quad b^{5} = \begin{bmatrix} 0 \\ 0 \\ 0 \\ 0 \end{bmatrix}
    \]
\end{itemize}

The network architecture is visualized below. Note that it is a fully-connected network (the edges are not shown).
\begin{figure}[h]
    \centering
    \includegraphics[width=0.6\textwidth]{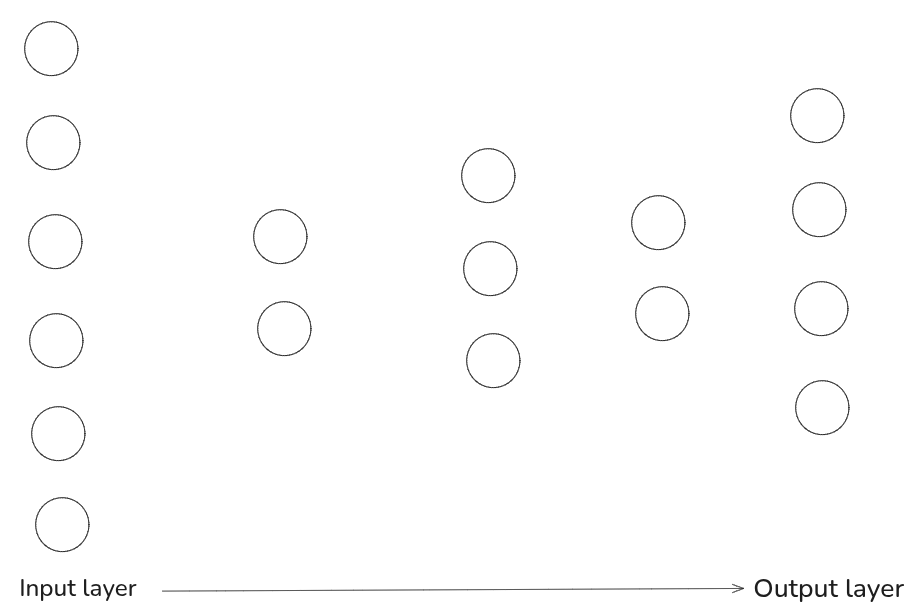}
    \caption{Neural network architecture for the uniform occlusion example.}
    \label{fig:nn2_example}
\end{figure}

\subsubsection{Analysis of Piecewise Behavior}

\paragraph{First Layer Pre-activations}
The pre-activations of the first layer, $z^{(1)}$, are affine functions of $c$:
\begin{align*}
z_1^1 &= (1\cdot 2) + (-1\cdot 3) + (-1\cdot c) + (1\cdot c) + (1\cdot c) + (1\cdot c) + 3 = 2c + 2 \\
z_2^1 &= (-1\cdot 2) + (1\cdot 3) + (-1\cdot c) + (1\cdot c) + (1\cdot c) + (1\cdot c) + 2 = 2c + 3
\end{align*}

\paragraph{Breakpoints}
The sign of the activation changes when $z_1^1 = 0$ or $z_2^1 = 0$. Solving for $c$:
\begin{itemize}
    \item $2c + 2 = 0 \implies c = -1$
    \item $2c + 3 = 0 \implies c = -3/2$
\end{itemize}
The set of breakpoints is $B_2 = \{-1.5, -1\}$. These values partition the real line into three intervals.

\paragraph{Layer Outputs by Interval}
The state of the network remains constant within each interval defined by the breakpoints. The last layer is ARGMAX of the $4^{th}$ hidden layer. As the $4^{th}$ layer is an affine layer, sign activation is not applied here.

\begin{table}[h]
\centering
\begin{tabular}{|c|c|c|c|}
\hline
\textbf{Interval} & $c < -1.5$ & $-1.5 \le c < -1$ & $-1 \le c$ \\
\hline
\textbf{Hidden Layer 1 Output} ($n_{1}^{2}$ $n_{2}^2$) & $\begin{bmatrix} -1 \\ -1 \end{bmatrix}$ & $\begin{bmatrix} 1 \\ -1 \end{bmatrix}$ & $\begin{bmatrix} 1 \\ 1 \end{bmatrix}$ \\
\hline
\textbf{Hidden Layer 2 Output} ($n_{1}^3$ $n_{2}^3$ $n_{3}^3$) & $\begin{bmatrix} -1 \\ 1 \\ 1 \end{bmatrix}$ & $\begin{bmatrix} 1 \\ 1 \\ 1 \end{bmatrix}$ & $\begin{bmatrix} 1 \\ 1 \\ -1 \end{bmatrix}$ \\
\hline
\textbf{Hidden Layer 3 Output} ($n_{1}^4$ $n_{2}^4$) & $\begin{bmatrix} 1 \\ 1 \end{bmatrix}$ & $\begin{bmatrix} 1 \\ 1 \end{bmatrix}$ & $\begin{bmatrix} 1 \\ -1 \end{bmatrix}$ \\
\hline
\textbf{Hidden Layer 4 Output} ($n_{1}^5$ $n_{2}^5$ $n_{3}^5$ $n_{4}^5$) & $\begin{bmatrix} 2 \\ 0 \\ 0 \\ -2 \end{bmatrix}$ & $\begin{bmatrix} 2 \\ 0 \\ 0 \\ -2 \end{bmatrix}$ & $\begin{bmatrix} 0 \\ -2 \\ 0 \\ 0 \end{bmatrix}$ \\
\hline
\end{tabular}
\caption{Network activations and outputs across the three stability intervals for the coloring constant $c$.}
\label{tab:piecewise_behavior}
\end{table}
\bibliographystyle{plain}
\bibliography{reference}
\end{document}